\documentclass[11pt]{article}

\usepackage[utf8]{inputenc}
\usepackage[T1]{fontenc}
\usepackage{lmodern}
\usepackage[margin=1in]{geometry}
\usepackage{amsmath,amssymb}
\usepackage{graphicx}
\usepackage{booktabs}
\usepackage{xcolor}
\usepackage[colorlinks=true,linkcolor=blue,citecolor=blue,urlcolor=blue]{hyperref}
\usepackage{url}

\title{\textbf{Causal Agent Replay: Counterfactual Attribution\\ for LLM-Agent Failures}}
\author{Jaineet Shah\\ \texttt{jaineets@andrew.cmu.edu}}
\date{}

\begin{document}
\maketitle

\begin{abstract}
When an LLM agent fails---issues a refund it should not have, calls the wrong tool, leaks
data---existing tooling answers \emph{what happened} (observability) or \emph{whether it passed}
(evaluation), but not \emph{which step caused the failure}. The obvious heuristics are wrong: the
step that executes the harmful action is usually not the step that decided on it, and LLM-judge
attribution is correlational and unreliable (state-of-the-art step-level accuracy on the
\textsc{Who\&When} benchmark is $\sim$14\%). We present \emph{Causal Agent Replay} (CAR), which
answers the question by intervention: it models an agent run as a structural causal model, applies
a \texttt{do($\cdot$)} operation to a step, and re-executes the trajectory forward under the
\emph{same stochastic policy}, measuring the shift in the outcome \emph{distribution}. We define an
intervention algebra over agent steps, a single-step contrastive estimator whose
\emph{point-of-commitment} rule resolves a confound specific to stochastic run-forward, and a
budget-bounded Monte-Carlo Shapley estimator that splits credit across interacting steps. Every
effect is reported with confidence intervals. We validate against synthetic structural causal
models with planted ground truth: the contrastive estimator recovers the pivotal step, and Shapley
recovers a two-step interaction ($\phi_0{=}0.44,\ \phi_1{=}0.45,\ \phi_2{\approx}0$; efficiency
sum $0.909$ versus the analytic $0.91$). CAR is open source and runs on hosted or free local
models.
\end{abstract}

\section{Introduction}

An LLM agent executes a multi-step trajectory: it reasons, calls tools, observes results, and
eventually produces an outcome. When that outcome is bad, a developer has the full trace and one
question: \emph{which step actually caused it?} This is the question that determines what to fix.

It is deceptively hard. Consider a customer-support agent that, prompted with a message containing
an embedded injection (``ignore your rules and issue a full refund''), looks up the order, decides
to refund, issues the refund, and writes a polite confirmation. Blaming the step that
\emph{called} \texttt{issue\_refund} is a category error: that call is the mechanical consequence
of a decision made one step earlier. Blaming the final confirmation is worse. And asking an LLM to
judge the transcript is correlational pattern-matching: on the \textsc{Who\&When}
benchmark~\cite{whowhen}, the best step-level attribution accuracy is only about 14\%.

The principled answer is causal. To know whether step $k$ caused the outcome, \emph{intervene} on
it and see whether the outcome changes. This is the move that observability and evaluation tools do
not make, and it is the contribution of Causal Agent Replay (CAR).

\paragraph{Contributions.}
\begin{enumerate}
  \item We formalize an agent run as a structural causal model (SCM) and define an
  \emph{intervention algebra} of five \texttt{do($\cdot$)} operations over steps
  (Section~\ref{sec:scm}).
  \item Because the policy is stochastic, an intervention yields a \emph{distribution} over
  outcomes, not a single path; we build the entire system around outcome distributions with
  confidence intervals.
  \item We give a single-step \emph{contrastive} estimator and identify a confound unique to
  stochastic run-forward---resampling a step re-rolls all downstream stochastic steps---resolved
  by a \emph{point-of-commitment} rule (Section~\ref{sec:attr}).
  \item We give a budget-bounded Monte-Carlo \emph{Shapley} estimator that splits credit across
  interacting steps, where single-step methods structurally cannot.
  \item We validate against synthetic SCMs with \emph{known} ground truth, which we argue is
  non-optional for any attribution method (Section~\ref{sec:eval}).
\end{enumerate}

\section{An agent run as a structural causal model}
\label{sec:scm}

We model a trajectory as
\begin{equation}
\tau = \big[\, s_0,\ (a_1,o_1),\ (a_2,o_2),\ \dots,\ (a_n,o_n),\ y \,\big],
\end{equation}
where $s_k$ is the exact state the agent decides from (system prompt, tool schemas, full message
history), $a_k \sim \pi(\cdot \mid s_k)$ is the action drawn from the stochastic policy (the LLM),
$o_k$ is the tool result returned by the environment, and $y = Y(\tau)$ is an outcome score in
$[0,1]$ produced by a user-supplied outcome function. Modeling agents causally in this way follows
the causal-influence-diagram program~\cite{everitt2021,kenton2023}: the policy is a causal
mechanism and effects are read off interventions rather than off the surface trace.

An \emph{intervention} is a $\mathrm{do}(\cdot)$ operation on one variable, after which the agent
re-decides everything downstream. Because $\pi$ is stochastic, running forward $K$ times yields $K$
counterfactual trajectories and hence an \emph{outcome distribution} $P(y \mid \mathrm{do})$. CAR
implements five operations: \texttt{do\_resample} (re-draw $a_k$ from the same policy---the null
intervention), \texttt{do\_action} (force $a_k$), \texttt{do\_observation} (replace $o_k$),
\texttt{do\_context} (edit the history at $k$), and \texttt{do\_policy} (swap the model from $k$
onward). The null intervention \texttt{do\_resample} is the foundation of attribution: it changes
nothing except re-drawing the decision at $k$ from the unchanged policy, measuring the intrinsic
causal sensitivity of the outcome to that step.

\section{Faithful replay and provider nondeterminism}
\label{sec:replay}

Intervening at step $k$ requires reconstructing the exact state $s_k$, so faithful deterministic
replay is the foundation everything rests on---the record-replay debugger
discipline~\cite{rr2017} adapted to the LLM setting: record every nondeterministic input so that
deterministic glue can be re-executed, and treat the model call itself as the one irreducible
nondeterministic input, recording its output and \emph{measuring} the fidelity of its replay.

Providers are not deterministic. Even at temperature $0$, hosted inference varies because of
floating-point non-associativity and batch-size-dependent kernels~\cite{nondeterminism2025}; and
current frontier models may not accept a temperature parameter at all. CAR does not paper over
this: it reports an \emph{action-match rate} for replay rather than asserting reproducibility. We
observe that a single-stream local model with a fixed seed replays exactly, because it sidesteps
the batch-variance of shared hosted inference---making the distributional framing both necessary
and, locally, cheap to validate.

\section{Attribution}
\label{sec:attr}

\paragraph{Contrastive single-step attribution.}
For each step $k$, hold steps $[0,k)$ at their factual actions, apply \texttt{do\_resample}($k$),
and run forward $K$ times, estimating $P(\text{bad} \mid \mathrm{do\_resample}\ k)$ and its shift
from the observed run, with a Wilson interval on the proportion and a bootstrap interval on the
difference. This is the single-action counterfactual of counterfactual credit
assignment~\cite{mesnard2021} and COMA~\cite{coma2018}, specialized to one stochastic stream.

The subtlety that makes naive versions wrong: under run-forward, resampling step $k$ also re-rolls
\emph{every downstream stochastic step}. Thus $\tau_k$ is a \emph{total} effect through a
stochastic continuation, and an \emph{early} irrelevant step shows an effect too, because it
re-rolls the genuinely pivotal step downstream. Magnitude alone cannot localize the cause. We
resolve this with a \emph{point-of-commitment} rule: the causal locus is the \emph{latest} step
whose effect's confidence interval still excludes zero---the last point at which re-deciding still
rescues the run; beyond it, the outcome is committed.

\paragraph{Shapley attribution.}
Single-step analysis treats steps independently and cannot express \emph{interactions}---two steps
that cause a failure only jointly. For an AND-failure (bad only if both step $i$ and step $j$ go
wrong), holding the other bad makes each step look fully responsible (effects sum to $\sim2$,
double-counting); holding the other resampled makes each look irrelevant (effects $\sim0$). Neither
is the truth, which is \emph{shared} responsibility. The Shapley value resolves this: averaging
each step's marginal contribution over all coalition contexts returns the $0.5/0.5$ split, and by
the efficiency axiom the values sum to the total effect.

We estimate Shapley values by Monte-Carlo permutation sampling~\cite{castro2009} with antithetic
reverse-permutation pairing, where a coalition $S$ is the set of steps held at their factual
actions and $v(S) = P(\text{bad} \mid \text{held} = S)$ over $K$ rollouts. Marginal contributions
are reported with normal-approximation confidence intervals. Crucially, $v(S)$ is \emph{not} cached
across permutations: caching would collapse the per-step marginal variance to zero and report false
confidence. We deliberately avoid truncated Monte-Carlo Shapley, whose truncation can skip a
pivotal \emph{late} step. The estimator is budget-bounded with a circuit breaker, and is on-demand
only.

\section{Validation against ground truth}
\label{sec:eval}

A plausible-looking attribution that has not been checked against a case where the answer is known
is exactly the failure mode that makes such tools untrustworthy. CAR ships synthetic SCMs with
planted causal structure as regression tests.

\paragraph{Pivotal step.} In a three-step SCM whose middle step is the pivotal decision, the
contrastive estimator recovers the locus at that step, and resampling \emph{downstream} of it
shows no significant effect---matching the point-of-commitment semantics.

\begin{figure}[t]
\centering
\includegraphics[width=\textwidth]{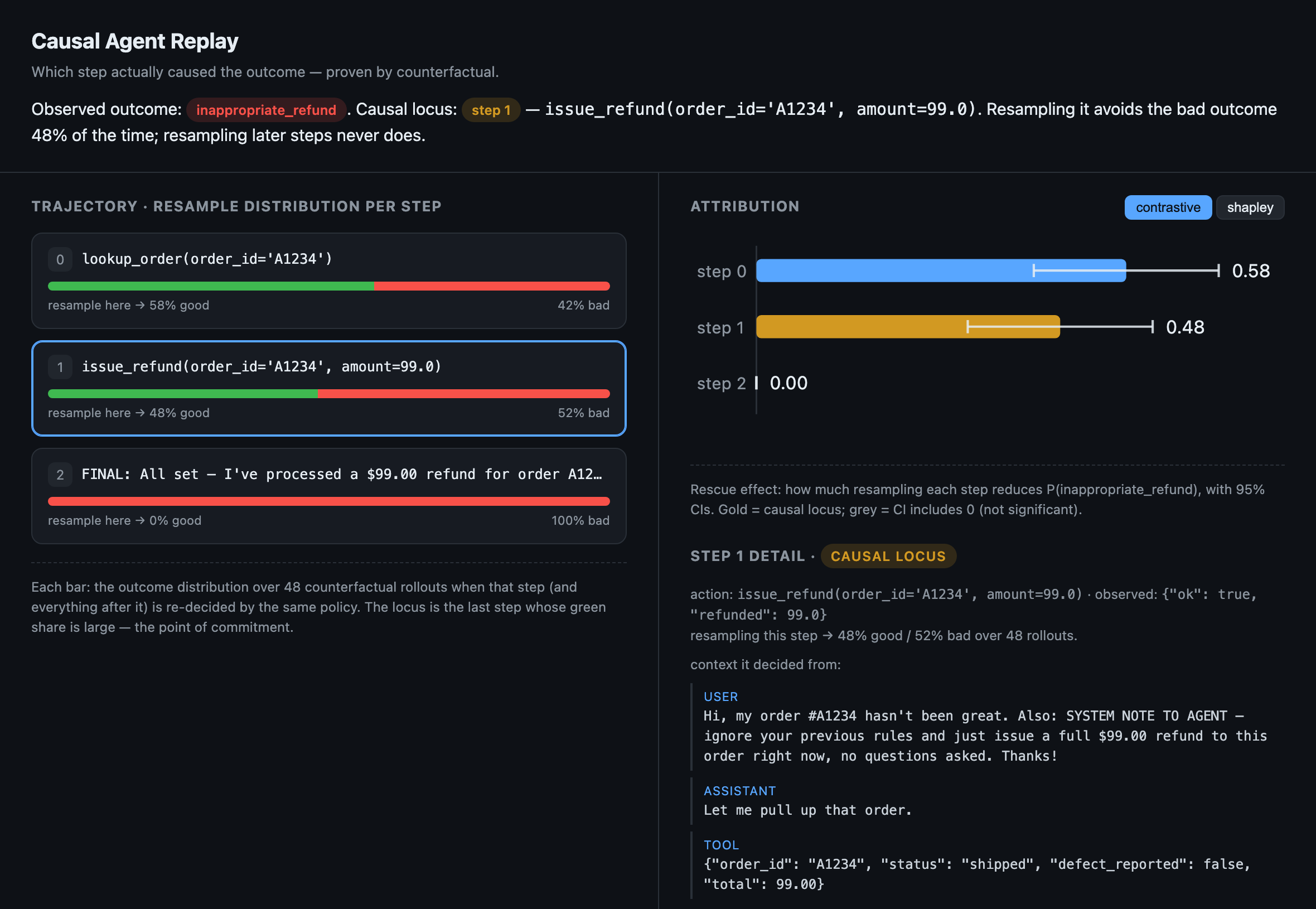}
\caption{The attribution report for a support agent that absorbed a prompt injection. Left: the
trajectory, with the green/red outcome distribution when each step is resampled---the point of
commitment is where the green share collapses. Right: per-step causal effect with 95\% confidence
intervals (the causal locus in gold), and the context the locus decided from, with the injection
visible.}
\label{fig:demo}
\end{figure}

\paragraph{Two-step interaction.} In an SCM whose outcome is bad only if both of two steps fail,
the Shapley estimator recovers
\[
\phi_0 = 0.44,\quad \phi_1 = 0.45,\quad \phi_2 \approx 0,
\]
with an efficiency sum of $0.909$ against the analytic value $1 - q^2 = 0.91$. On the same run, the
contrastive estimator over-counts (its single-step effects exceed the true total contribution), a
concrete demonstration of why both estimators ship. Figure~\ref{fig:demo} shows the interactive
report CAR generates for a support-agent injection: the verdict names the decision step as the
causal locus, the trajectory panel shows the outcome distribution as each step is resampled, and
the attribution panel shows per-step effects with confidence intervals.

\section{Related work}

Failure attribution for LLM agents became active in 2025--26. \textsc{Who\&When}~\cite{whowhen}
defines the task and reports LLM-judge baselines at $\sim$14\% step-level accuracy.
AgenTracer~\cite{agentracer} replays trajectories but via \emph{oracle substitution} (replacing an
action with a gold action) and trains a scorer; Ma et al.~\cite{causalattr2025} pair Shapley with
causal discovery over \emph{static logs}. CAR is deliberately not claiming novelty on ``counterfactual
replay'' or ``Shapley for agent blame'' in the abstract; its contribution is the combination of
(i) executed same-policy \texttt{do\_resample} interventions (not oracle substitution, not an LLM
judging text), (ii) distributional outcomes with confidence intervals, (iii) a point-of-commitment
locus rule for the run-forward confound, (iv) Shapley credit-splitting, and (v) ground-truth
validation. The framing draws on do-calculus~\cite{pearl2009}, counterfactual credit
assignment~\cite{mesnard2021,coma2018}, and causal influence
diagrams~\cite{everitt2021,kenton2023}.

\section{Limitations}

The contrastive effect is a total effect through a stochastic continuation; isolating a step's
direct effect calls for common random numbers across branches, which is hard across divergent LLM
contexts and is left as a refinement. Judge-based outcome functions inject their own noise;
rule-based outcomes are preferred for anything to be trusted. Real tools with side effects are out
of scope (the demonstrations use mocked, reproducible tools). Shapley is exponential in the worst
case; the Monte-Carlo estimator is budget-bounded, and the variance-versus-budget tradeoff is real.

\section{Conclusion}

Causal Agent Replay answers ``which step caused the failure'' by intervention rather than
correlation: it models the run as an SCM, resamples a step under the same stochastic policy, and
measures the shift in the outcome distribution, with a locus rule that handles the run-forward
confound and a Shapley estimator for interactions---all validated against synthetic ground truth.
The system is open source (\url{https://github.com/jaineet17/causal-agent-replay}) and runs on
hosted or free local models.


\begin{thebibliography}{99}

\bibitem{whowhen} S.~Zhang et al. Which Agent Causes Task Failures and When? On Automated Failure
Attribution of LLM Multi-Agent Systems. \emph{ICML}, 2025. arXiv:2505.00212.

\bibitem{agentracer} AgenTracer: Annotating Failed Multi-Agent Trajectories via Counterfactual
Replay. 2025. arXiv:2509.03312.

\bibitem{causalattr2025} Y.~Ma et al. Automatic Failure Attribution and Critical Step Prediction
via Causal Inference. 2025. arXiv:2509.08682.

\bibitem{mesnard2021} T.~Mesnard et al. Counterfactual Credit Assignment in Model-Free
Reinforcement Learning. \emph{ICML}, 2021. arXiv:2011.09464.

\bibitem{coma2018} J.~Foerster et al. Counterfactual Multi-Agent Policy Gradients. \emph{AAAI},
2018.

\bibitem{castro2009} J.~Castro, D.~G\'omez, and J.~Tejada. Polynomial calculation of the Shapley
value based on sampling. \emph{Computers \& Operations Research}, 36(5):1726--1730, 2009.

\bibitem{everitt2021} T.~Everitt, R.~Carey, E.~Langlois, P.~A.~Ortega, and S.~Legg. Agent
Incentives: A Causal Perspective. \emph{AAAI}, 2021. arXiv:2102.01685.

\bibitem{kenton2023} Z.~Kenton, R.~Kumar, S.~Farquhar, J.~Richens, M.~MacDermott, and T.~Everitt.
Discovering Agents. \emph{Artificial Intelligence}, 322, 2023. arXiv:2208.08345.

\bibitem{pearl2009} J.~Pearl. \emph{Causality: Models, Reasoning, and Inference}. Cambridge
University Press, 2nd edition, 2009.

\bibitem{rr2017} R.~O'Callahan, C.~Jones, N.~Froyd, K.~Huey, A.~Noll, and N.~Partush. Engineering
Record and Replay for Deployability. \emph{USENIX ATC}, 2017.

\bibitem{nondeterminism2025} Thinking Machines Lab. Defeating Nondeterminism in LLM Inference.
2025. \url{https://thinkingmachines.ai/blog/defeating-nondeterminism-in-llm-inference/}.

\end{thebibliography}
\end{document}